\documentclass[letterpaper, 10 pt, conference]{ieeeconf}
\usepackage{times}
\usepackage[pdftex]{graphicx}
\usepackage{amsmath,amssymb,amsopn,amstext,amsfonts}
\usepackage{cancel}
\usepackage[space]{cite}
\usepackage{pdfsync}
\usepackage{balance}
\usepackage{color}
\usepackage{mathtools}
\usepackage{algorithm}
\usepackage{algorithmic}
\usepackage{bm}

\usepackage{diagbox}
\usepackage{float}
\usepackage{epstopdf}
\usepackage{url}
\usepackage{booktabs}
\usepackage{stfloats}
\usepackage{graphicx}
\usepackage{tablefootnote}
\usepackage{titlesec}

\usepackage[caption=false,font=normalsize,labelfont=sf,textfont=sf]{subfig}

\usepackage{verbatim}
\usepackage{fancyhdr}
\usepackage{subfig}
\usepackage{tikz}
\usepackage{hyperref}

\graphicspath{{./Figures/}}
\DeclareGraphicsExtensions{.pdf,.png,.jpg,.eps,.svg}

\IEEEoverridecommandlockouts
\title{\LARGE \bf NDOB-Based Control of a UAV with Delta-Arm Considering Manipulator Dynamics
}

\author{Hongming Chen$^{1}$, Biyu Ye$^{1}$, Xianqi Liang$^{1}$, Weiliang Deng$^{1}$  Ximin Lyu$^{1}$
    \thanks{$^{1}$All authors are with the School of Intelligent Systems Engineering, Sun Yat-sen University, Guangzhou, China.(\textit{Corresponding author:} Ximin Lyu)}%
    \thanks{E-mail: {\tt\small lvxm6@mail.sysu.edu.cn}}
    \thanks{This work was supported by the National Key R\&D Program of China under grant no. 2023YFB4706600} 
}

\begin{document}  
\maketitle

\thispagestyle{empty}
\pagestyle{empty}

\begin{abstract}
Aerial Manipulators (AMs) provide a versatile platform for various applications, including 3D printing, architecture, and aerial grasping missions.  However, their operational speed is often sacrificed to uphold precision. Existing control strategies for AMs often regard the manipulator as a disturbance and employ robust control methods to mitigate its influence.
This research focuses on elevating the precision of the end-effector and enhancing the agility of aerial manipulator movements. We present a composite control scheme to address these challenges. Initially, a Nonlinear Disturbance Observer (NDOB) is utilized to compensate for internal coupling effects and external disturbances. Subsequently, manipulator dynamics are processed through a high pass filter to facilitate agile movements.
By integrating the proposed control method into a fully autonomous delta-arm-based AM system, we substantiate the controller's efficacy through extensive real-world experiments. The outcomes illustrate that the end-effector can achieve accuracy at the millimeter level.
\end{abstract}

\vspace{-1pt}
\section{INTRODUCTION}
\vspace{-2pt}
In recent years, Aerial Manipulators (AMs) have emerged as a significant research focus in the field of aerial robotics, owing to their versatility and larger workspace. They are widely used in various scenarios, including 3D printing~\cite{zhang2022aerial}, inspection~\cite{trujillo2019novel}, and aerial grasping~\cite{zhang2018grasp}. Integrating a dexterous manipulator into an Unmanned Aerial Vehicle (UAV) introduces additional degrees of freedom (DoF) and expands the workspace. This inherent redundancy can be effectively leveraged by assigning multiple subtasks, thereby enabling AMs to perform complex operations~\cite{ruggiero2018aerial}.

Despite their potential, current control strategies often treat the manipulator as a disturbance and employ robust control methods to mitigate its influence~\cite{aydemir2020evaluation,chen2022adaptive,jiao2020disturbance}. 
These approaches come with significant limitations. More specifically, in the context of an AM system, the Nonlinear Disturbance Observer (NDOB) is effective at compensating for low-frequency disturbances~\cite{chen2015disturbance}. However, when the manipulator undergoes high-speed motion, it generates rapidly changing reactive forces that are exerted on the floating base and can be characterized as high-frequency signals. This necessitates a slower operational speed to ensure the NDOB can adequately address these disturbances~\cite{khalifa2016hybrid}. Consequently, the primary drawback of current NDOB-based approaches is their inability to handle the high-frequency dynamics associated with fast manipulator movements, which limits the overall performance and agility of the system.

\begin{figure}
     \vspace{16pt}
    \centering
    \includegraphics[width=1\linewidth]{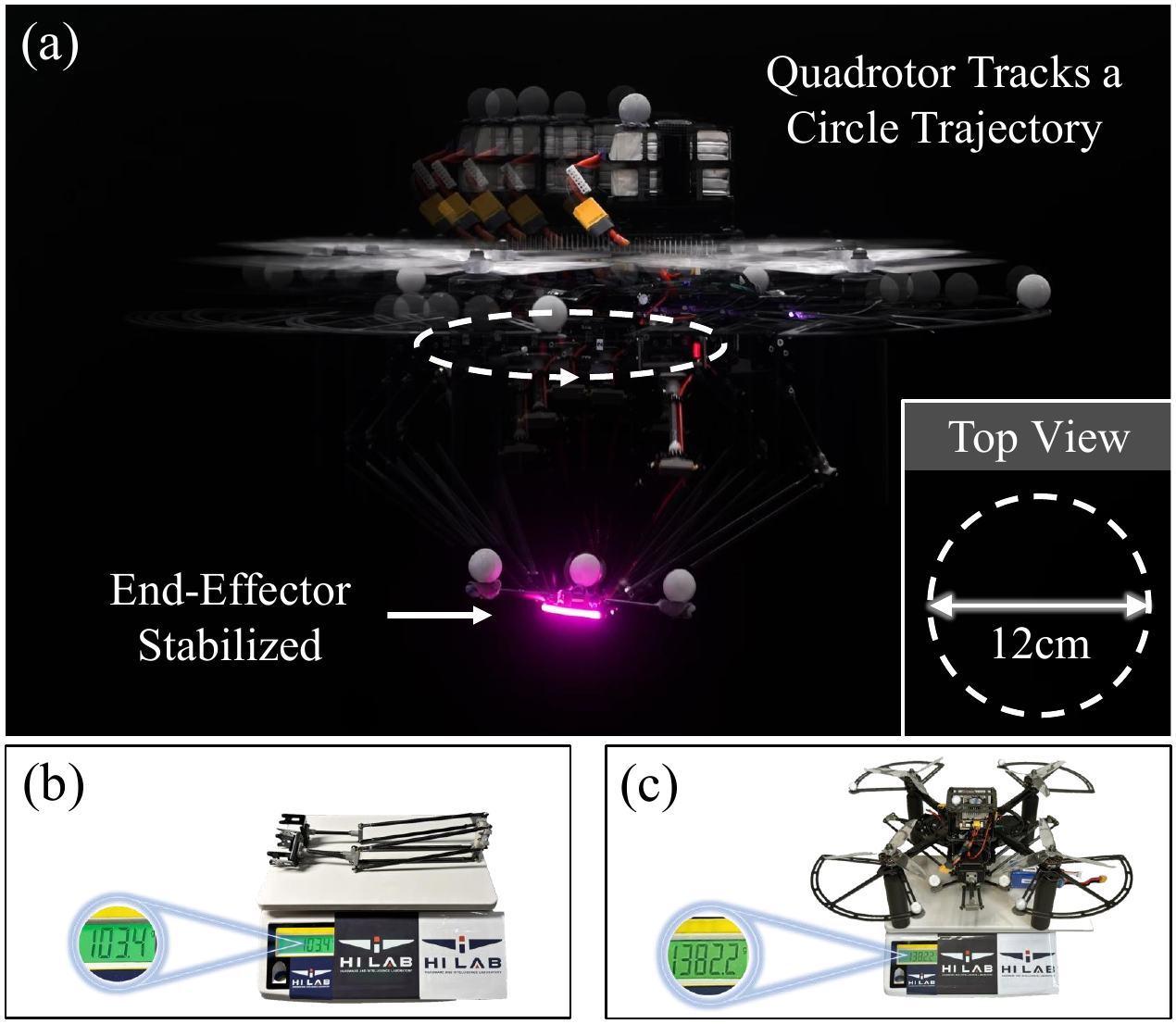}
    \caption{(a) The UAV performs a circular maneuver while keeping the end-effector stabilized at the origin. The dotted line represents the UAV trajectory, and the position error of the end-effector is only 6 mm. (b) The Delta-arm weighs 103.4 g. (c) The aerial manipulator platform, designed for various tasks, weighs 1382.2 g.}
    \label{fig:1}
    \vspace{-20pt}
\end{figure}

To address the aforementioned challenges, this paper proposes a composite control framework. This framework utilizes an NDOB to compensate for low-frequency disturbances and employs end-effector dynamics, processed through a high pass filter, to mitigate high-frequency disturbances.
Implemented with the proposed control framework, the quadrotor can effectively manage the reactive forces generated by the high-speed manipulator. 
Even when the end-effector carries an additional payload~(400~g, representing 30\% of the AM's weight) and performs high-speed motions~(end-effector max acceleration achieves 0.3g), we can maintain a positioning error within 2.9 cm for the quadrotor.
Furthermore, we achieve a tracking accuracy of 6 mm for the end-effector while the UAV maintains agile movement(\textit{e.g.}, end-effector stabilization shown in Fig.~\ref{fig:1}). 
We summarize our contributions as follows:
\begin{itemize}
    \item We propose a composite control framework for the AM platform, incorporating an NDOB and dynamic high-frequency compensation for the manipulator. This framework effectively mitigates the effect of the manipulator's agile motions, facilitating the end-effector to attain millimeter-level precision even during high-speed operations.
    \item The proposed system can control the AM in windy environments, achieving a trajectory tracking error of 6 mm for the manipulator, which compensates for the UAV trajectory tracking error. Such conditions are commonly encountered during fieldwork.
    \item We integrate our proposed AM control method into an AM system and validate its performance through extensive real-world experiments. The results of these experiments provide strong evidence supporting the effectiveness and reliability of our control framework.
\end{itemize}

This paper is organized as follows: Section \ref{SECTION II} provides an overview of the related work on aerial manipulators. Section \ref{SECTION III} presents the kinematic and dynamic model of the aerial manipulator. In Section \ref{SECTION IV}, we detail the proposed control method. Section \ref{SECTION V} demonstrates the experimental results. Finally, Section \ref{SECTION VI} concludes the paper.

\section{RELATED WORK}
\label{SECTION II}
\subsection{Overview of AMs Types }
The AMs system inherently exhibits nonlinearity, presenting a considerable control challenge, especially during high-speed movements of the end-effector with a heavy payload. Previous studies such as \cite{kim2018stabilizing,ali2020controlling,jafarinasab2019model} have predominantly focused on quadrotor-mounted serial manipulators. While serial manipulators are more prevalent and easier to analyze, they suffer from a notable drawback in aerial transportation. 
For example, they require increased torque when handling equivalent weight payloads compared to parallel manipulators. Additionally, serial manipulators find it more difficult to achieve the same accuracy as parallel manipulators do~\cite{tsai1999robot}. 

Recent works \cite{zhang2022aerial,danko2015parallel,bodie2021dynamic} have addressed the challenges of parallel manipulators. These approaches allow robots to achieve precise positions, surpassing the payload limitations of serial manipulators. Zhang . al.~\cite{zhang2022aerial} prioritize precision control in the 3D printing process, employing inverse kinematics for position regulation and utilizing Nonlinear Model Predictive Control (NMPC) to control the aerial manipulators. However, in the 3D printing process, the manipulator operates at a relatively slow speed. Bodie \textit{et al.}~\cite{bodie2021dynamic} initially propose leveraging the dynamics of the end-effector. However, they do not integrate the payload into their system to fully demonstrate its effectiveness.

\subsection{Control Approaches for AMs}
Dynamic coupling presents a critical control challenge in aerial manipulators. Conventional aerial manipulator control methods typically use a DOB to robustly control the floating base without accounting for the dynamics of the manipulator~\cite{khalifa2016hybrid}. Current control approaches can be classified into two main categories: full-body control and separate control. Full-body control involves simultaneous regulation of both the flying platform and the manipulator within the system \cite{lippiello2012cartesian}. This approach requires a precise nonlinear model to ensure control accuracy and robustness. However, obtaining such a precise nonlinear model can be challenging. On the other hand, separate control considers the manipulator's reacted wrench as a disturbance and does not require considering the entire system, making it easier to implement. 

For full-body control, Lunni \textit{et al.}~\cite{lunni2017nonlinear} propose an NMPC method that considers the full-body dynamics. They address the collaborative planning problem of the AMs by formulating it as a constrained optimization problem. 
Lee \textit{et al.}~\cite{lee2020aerial} utilize Model Predictive Control (MPC) as a sub-optimal planner to generate trajectories. Dimos \textit{et al.}~\cite{tzoumanikas2020aerial} propose a model that considers only static forces, neglecting the impact of the manipulator on the quadrotor. 
Their approach focuses solely on the effect of contact forces on the system.  However, this simplification of the system overlooks the manipulator and leads to degraded position tracking.

For separate control, Cao \textit{et al.}~\cite{cao2023eso} consider the manipulator's wrench as a disturbance factor while overlooking external disturbances like wind. Zhang \textit{et al.}~\cite{zhang2019robust} incorporate model-based disturbance estimators into the controllers to compensate for dynamic coupling effects, which requires precise modeling of the aerial manipulators. Wang \textit{et al.}~\cite{wang2024precise} employ a neural network to estimate the composite dynamic model. However, data collection remains challenging. Jimenez \textit{et al.}~\cite{jimenez2013control} address external disturbances in nonlinear control using backstepping control. However, this method is sensitive to external disturbances and measurement noise.

\begin{figure}
    \centering
    \includegraphics[width=1\linewidth]{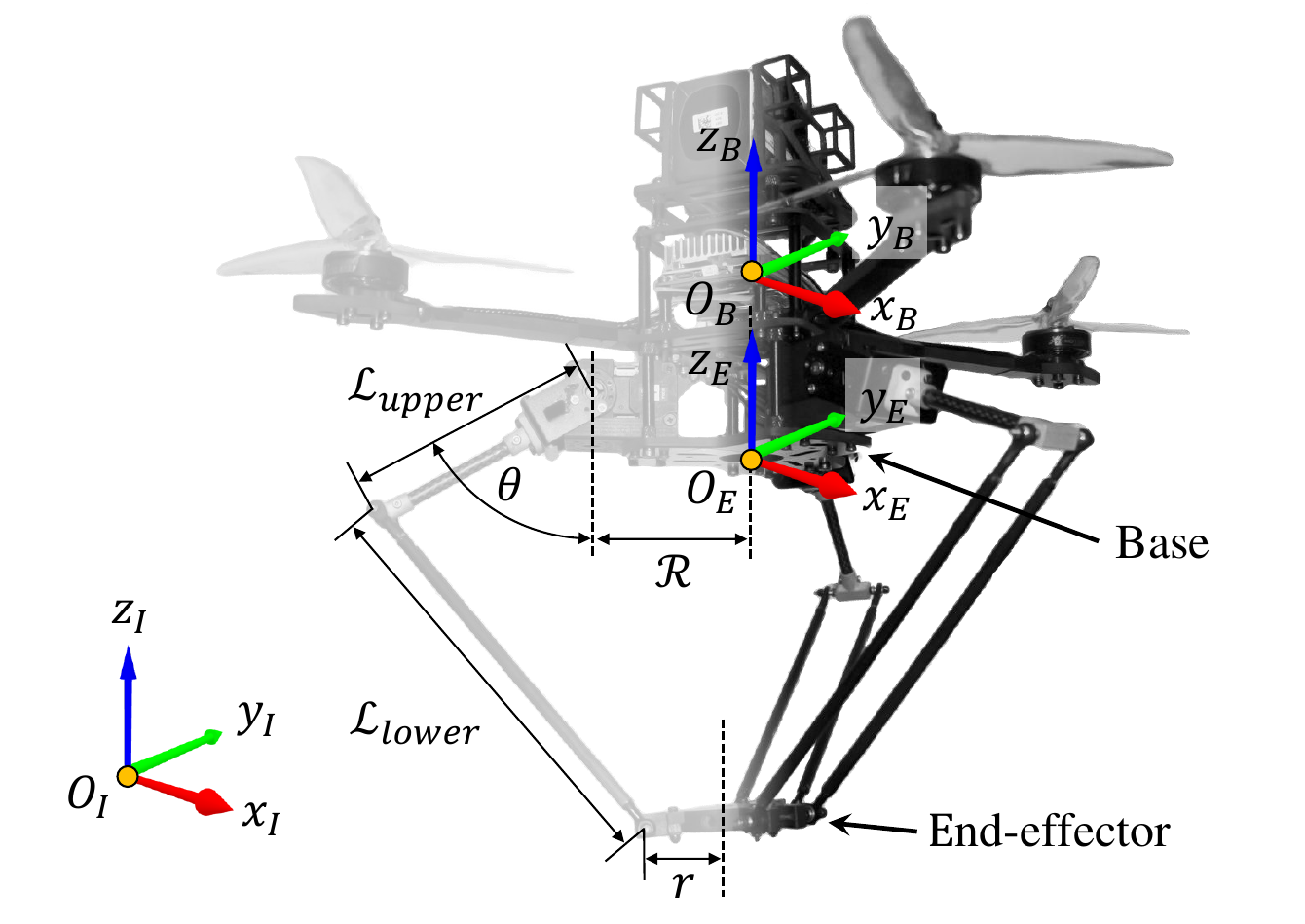}
    \caption{Schematics of the delta parallel robot and the coordinate frames of the aerial manipulator (AM) system.}
    \label{fig:coordinate}
    \vspace{-25pt}
\end{figure}

\section{MODELING}
\label{SECTION III}

Throughout this article, three right-handed coordinate frames are utilized. These are the inertial frame $\mathcal{F}_I:\{\bm{x}_I,\bm{y}_I,\bm{z}_I\}$, the quadrotor body-fixed frame $\mathcal{F}_B:\{\bm{x}_B,\bm{y}_B,\\\bm{z}_B\}$, and the end-effector frame $\mathcal{F}_E:\{\bm{x}_E,\bm{y}_E,\bm{z}_E\}$. We use bold lowercase letters to represent vectors and bold uppercase letters for matrices. Vectors with subscription B and E are expressed in the body-fixed frame and end-effector frame, respectively. Otherwise, it is in the inertial frame.  (see Fig.~\ref{fig:coordinate}).

\subsection{Dynamics of the Aerial Manipulator}
The dynamics of the aerial manipulator comprise two components. The first component pertains to the quadrotor and the second component involves the delta-arm reactive term. Based on~\cite{cao2023eso}, 
the dynamics of the quadrotor in the aerial manipulator system are:
\begin{equation}
\begin{split}
\Ddot{\bm{p}} &= \frac{T\bm{z}_B - \bm{f}^e }{m_B+m_M}  +\bm{g} \\
\bm{\tau} &= \mathcal{I}\cdot{\dot{\bm{\omega}}}+ \bm{\omega}\times (\mathcal{I}\bm{\omega}) - \bm{\tau}^e \\ 
\dot{\bm{R}}&=\bm{R}\hat{\bm{\omega}}
\label{con:equ1}
\end{split}
\end{equation}
where $T$, $m_B$, and $m_M$ represent the collective thrust,  the quadrotor mass, and the manipulator mass, respectively.
 $\bm{z}_B$ denotes the z-axis of the body frame in the inertial frame. $\mathcal{I}$ and $\bm{\omega}$ represent the inertia matrix and the angular velocity vector of the quadrotor, respectively. $\bm{f}^e$ and $\bm{\tau}^e$ represent the force and torque caused by external disturbance. 

When the manipulator operates at high speed, directly modeling 
the coupling terms become challenging due to the variability of the inertia matrix and the resulting reaction torques. However, given the lightweight nature of the delta-arm, which constitutes only 7.5\% of the aerial manipulator system, the delta-arm's coupling term is relatively small.
In this paper, we categorize the external wrench into two components: the low-frequency part and the high-frequency part~\cite{kim2018discrete}, which can be represented: 

\begin{align}
\bm{f}^e &= \bm{f}^{high} + \bm{f}^{low}\label{con:equ13a}\\
\bm{\tau}^e &= \bm{\tau}^{high} + \bm{\tau}^{low}
\label{con:equ13b}
\end{align}

The low-frequency segment encompasses wind effects, the delta-arm's coupling term, and the low-frequency part of the wrench produced by the end-effector, whereas the high-frequency segment includes the high-frequency part of the wrench produced by the end-effector. The low-frequency external wrench and high-frequency external wrench are, respectively, handled by the low-frequency disturbance estimator and high-frequency disturbance estimator, which are detailed in Sections~\ref{sec: nonlinear disturbance observer} and Sections~\ref{sec:end-effector dynamic compensate}.

\subsection{Position and Velocity Kinematics of delta-arm}
\label{sec:Position and Velocity Kinematics of the delta-arm}
There are various methods available for analyzing delta-arm position kinematics. In this study, we adopt the approach outlined in \cite{codourey1996dynamic}. The position kinematics can be represented as: 
\begin{equation}
\left\|\bm{p}_E^e-\bm{h}_i\right\|^2=\mathcal{L}_{lower}^2,\quad i=1,2,3,
\label{con:velocity con1}
\end{equation}
where 
\begin{equation}
\bm{h}_i=\biggl[\begin{array}{c}-(\mathcal{R}-r+\mathcal{L}_{upper}\cos q_i)\cos[(i-1)\pi/3]\\(\mathcal{R}-r+\mathcal{L}_{upper}\cos q_i)\sin[(i-1)\pi/3]\\\mathcal{L}_{upper}\sin q_i\end{array}\biggr],
\label{con:velocity con2}
\end{equation}
$\bm{p}_E^e$ represents the position of end-effector in $\mathcal{F}_E$ frame. As illustrated in Fig.~\ref{fig:coordinate}. $\mathcal{L}_{upper}\in \mathbb{R}$ and $\mathcal{L}_{lower}\in \mathbb{R}$ represent the lengths for upper arms and lower arms
, respectively.  
$\mathcal{R}$ and $r$ represent the circumcircle radius of the static platform and end-effector, respectively. $\bm{q} = [ q_1, q_2, q_3 ]^T \in \mathbb{R}^3$  represents the servo angle vector.

Utilizing forward kinematics, the position $\bm{p}_E^e$ can be determined from~\eqref{con:velocity con2} with a given joint vector $\bm{q}$. Subsequently, leveraging these computations, prior studies~\cite{hoai2023design} have controlled the manipulator and positioned it as required.
However, relying solely on position kinematics proves insufficient for effective control of the delta-arm. Using position kinematics alone allows only for intermittent precise positioning. It is more intuitive to prioritize the control of speed and other low-level aspects. Given the characteristics of parallel manipulators, earlier research \cite{carp2003dynamic} iteratively determines velocity.
To address the aforementioned problem, we adopt the closed form velocity for delta-arm proposed by \cite{guglielmetti1994closed}, defining:
\begin{equation}
\bm{p}_{Ei}^{e} =\bm{\phi}_i \cdot \bm{p}_E^e, 
\label{con:velocity con3}
\end{equation}
where  
\begin{equation}
     \quad  \bm{\phi}_i=\begin{bmatrix}\cos[(i-1)\pi/3]&\sin[(i-1)\pi/3]&0\\-\sin[(i-1)\pi/3]&\cos[(i-1)\pi/3]&0\\0&0&1\end{bmatrix}. 
     \label{con:velocity con4}
\end{equation}

Based on the structural characteristics of the delta-arm: 
\begin{equation}
\bm{\beta}_i = \frac{1}{\mathcal{L}_{lower}}(\bm{p}_{Ei}^e - \bm{e}_i ),
\label{con:velocity con5}
\end{equation}
where $\bm{e}_i$ is the $i^{th}$ elbow in the $(O,x_i,z)$ frame with:
\begin{equation}
\begin{split}
\bm{e}_i = \begin{bmatrix}{\mathcal{R}-r}\\0\\0\end{bmatrix}+\mathcal{L}_{upper}\bm{\alpha}_\mathrm{i},\quad
\bm{\alpha}_i = \begin{bmatrix}\cos{q_i}\\0\\\-sin{q_i}\end{bmatrix}.
\label{con:velocity con con6}
\end{split} 
\end{equation}

Apply a transformation to Equation~\eqref{con:velocity con5} and differentiate both sides:
\begin{equation}
\begin{split}
\dot{\bm{p}_{Ei}^e} = \mathcal{L}_{upper}\cdot\dot{\bm{\alpha}_i}\cdot\dot{q_i}
+ \mathcal{L}_{lower}\cdot\dot{\bm{\beta}_i}.
\label{con:velocity con7}
\end{split}
\end{equation}

we can get the compact form:
\begin{equation}
\begin{split}
\bm{M}\cdot \dot{\bm{p}_E^e} =   \mathcal{L}_{upper}\cdot\bm{V}\cdot{\dot{\bm{q}}}
\label{con:velocity con8},
\end{split}
\end{equation}
where 
\begin{figure*}
    \centering
    \includegraphics[width=1\linewidth]{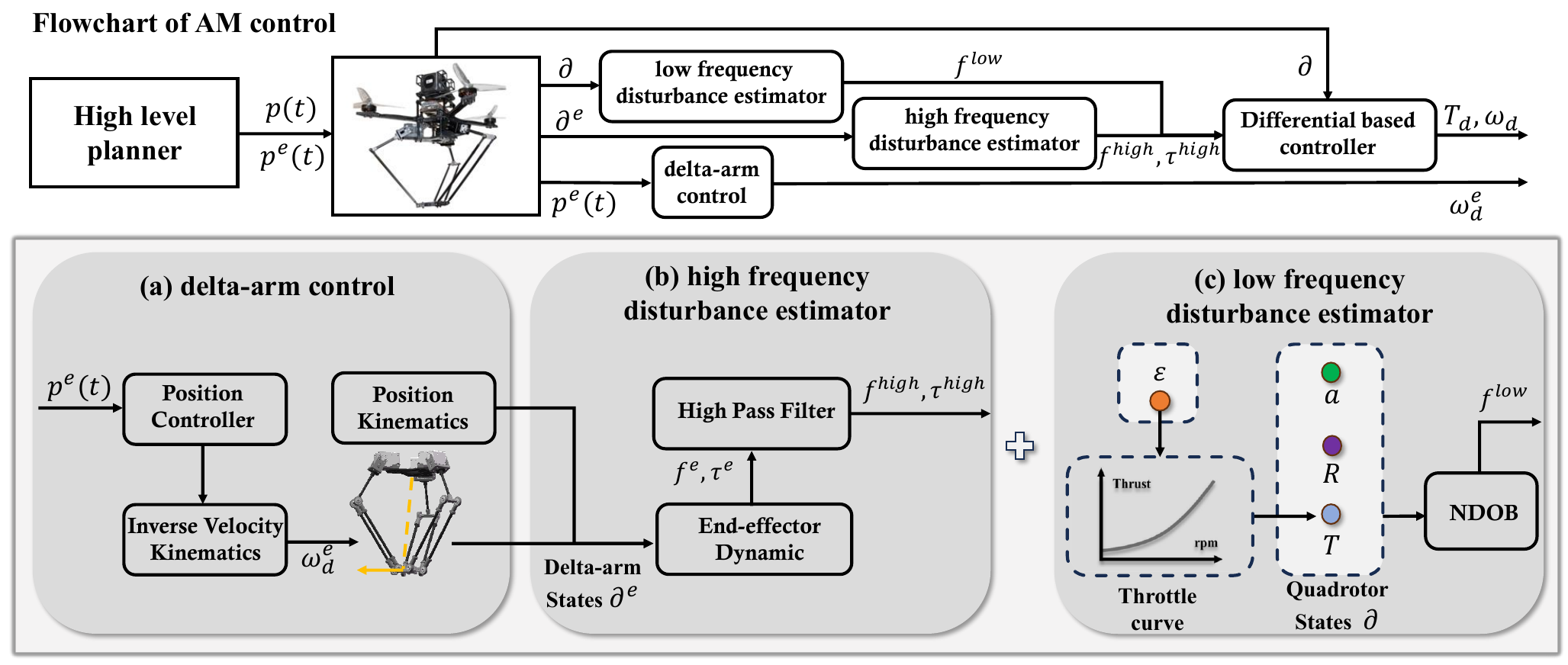}
    \caption{Flowchart of our control framework: The desired quadrotor and delta-arm trajectory passes through three blocks in sequence to obtain the servo motor desired agile velocity $\omega_d^e$, the AM desired thrust $T_d$, and the AM desired angular velocity $\omega_d$.
    \textbf{(a)} delta-arm control: process the delta-arm desired trajectory $p^e(t)$ to obtain the servo desired angular velocity. 
    \textbf{(b)} High-Frequency Disturbance Estimator: The end-effector dynamics pass through a high pass filter to obtain $f^{high}$.
    \textbf{(c)} Low-Frequency Disturbance Estimator: The actual state $\partial$ of the quadrotor passes through an NDOB to estimate $f^{low}$.
    } 
    \label{fig:3}
     \vspace{-10pt}
\end{figure*}
\begin{equation}
\bm{M} = 
\left[
\begin{smallmatrix}
\bm{\beta}_1^T \cdot \bm{\phi}_1 \\
\bm{\beta}_2^T \cdot \bm{\phi}_2 \\
\bm{\beta}_3^T \cdot \bm{\phi}_3 
\end{smallmatrix}
\right]
,
\;
\bm{V} = 
\left[
\begin{smallmatrix}
\bm{\beta}_1^T \cdot \dot{\bm{\alpha}}_1 & 0 & 0 \\
0 & \bm{\beta}_2^T \cdot \dot{\bm{\alpha}}_2 & 0 \\
0 & 0 & \bm{\beta}_3^T \cdot \dot{\bm{\alpha}}_3
\end{smallmatrix}
\right].
\label{con:velocity con9}
\end{equation}

By employing the expression \eqref{con:velocity con8}, we can elucidate the velocity kinematics of the end-effector, denoted as $\dot{\bm{p}}_{E}^e$. 
The position of the end-effector in the inertial frame is:
\begin{equation}
\begin{split}
\vspace{-10pt}
\bm{p}^e= \bm{p}^b+\bm{R}^E\bm{p}^e_E,
\label{con:velocity con10}
\end{split}
\end{equation}
where $\bm{p}^b$ is the position of the link base.  The time derivative of $\bm{p}^e$ is:
\begin{equation}
\begin{split}
\dot{\bm{p}}^e= \dot{\bm{p}}^b+\bm{\omega}\times\bm{R}^E\bm{p}_E^e+\bm{R}^E\dot{p}_E^e,
\label{con:velocity con11}
\end{split}
\end{equation}
where all the parameters are known, allowing for precise control of the desired trajectory. 



\section{CONTROLLER DESIGN}
\label{SECTION IV}
\label{controller design}

With the kinematic and dynamic analysis of the AM system completed, the subsequent step involves designing the framework. The flowchart of control framework is depicted in Fig~\ref{fig:3}. The higher-level planner autonomously generates a dynamically flexible trajectory for both the quadrotor and the delta-arm~\cite{MelKum1105}. Subsequently, for the quadrotor: a differential flatness-based controller meticulously tracks this generated trajectory while incorporating a hierarchical disturbance compensation strategy. Specifically, this strategy consists of two main components: firstly, it employs an NDOB to counteract low-frequency external forces; secondly, it addresses high-frequency components through end-effector dynamics coupled with a high pass filter. For the delta-arm, a position controller is proposed. It utilizes the inverse velocity kinematics as a priori knowledge, which enables fast convergence.

\subsection{Nonlinear Disturbance Observer }
\label{sec: nonlinear disturbance observer}
To enhance robustness without incorporating an additional sensor, we utilize the quadrotor states $\partial$ to estimate the external disturbance (\textit{i.e.}, force). We adopt the NDOB proposed by Yu \textit{et al.}\cite{yu2024dobbasedwindestimationuav}. The article assumes the external disturbance to be constant with the force NDOB denoted as $\bm{f}^{low}$. By employing this assumption and its associated observer:
\begin{equation}
\begin{split}
\dot{\bm{f}}^{low}=\frac {c}{m}(m\bm{a}-mg\bm{e_}3+\bm{T}\bm{R}\bm{e}_3-\bm{f}^{low})
\label{con:equ22}
\end{split}
\end{equation}
and convert it to discretization
\begin{equation}
\begin{split}
\bm{f}^{low}(k+1) = {} & \bm{f}^{low}(k)+\frac{c}{m}(m\bm{a}-{} \\
                       & mg\bm{e}_3+\bm{T}\bm{R}\bm{e}_3-\bm{f}^{low})\Delta t.
\label{con:equ23}
\end{split}
\end{equation}

The parameter $c$ governs the convergence speed. $\bm{a}$ represents the acceleration derived from the IMU of the flight controller. $\bm{e}_3$ represents quadrotor thrust direction in the inertial frame, and $\bm{R}$ represents the rotation matrix from the inertial frame to the body frame. $\Delta t$ denotes the reciprocal of the sensor frequency. To ensure accurate utilization of the observer for estimating external wrench, system identification becomes imperative. This process involves identifying critical parameters such as the rotor motor thrust coefficient and the moment of inertia matrix. In an implementation, $\bm{f}^{low}$ is filtered with a second-order Butterworth filter at the cut-off frequency (50Hz) to ensure consistent phase and delay. 

\subsection{End-effector dynamic compensate}
\label{sec:end-effector dynamic compensate}
From (\ref{con:velocity con11}), we can easily obtain the end-effector velocity in closed form. By differentiating this velocity, we derive the end-effector acceleration. Consequently, we can determine the impact of the end-effector on the quadrotor as follows:
\begin{equation}
\begin{split}
\bm{f}^{end} &= \bm{R}^E\ddot{\bm{p}}_E^e\cdot m_{payload}
\\
\bm{f}^{high} &= F_h(\bm{f}^e)
\\
\bm{\tau}^{high} &= \bm{f}^{high}\times \bm{p}_E^{high}.
\label{con:equ16}
\end{split}
\end{equation}

The terms $\bm{f}^{end}$, $\bm{f}^{high}$, and $F_h$ represent the dynamic forces caused by the end-effector, the high-frequency dynamic term, and the high pass filter, respectively. To address these dynamics, $\bm{f}^{end}$ are passed through a high pass filter. This approach is justified for two reasons: first, there is excessive noise when directly differentiating the end-effector velocity; second, the NDOB has already compensated for the low-frequency components of the end-effector dynamics.

\subsection{Quadrotor Control Framework}
\label{quadrotor control framework}
\subsubsection{Position and Velocity Control}
For quadrotor, the controller receives the desired position ($\bm{p}_d$), velocity ($\bm{v}_d$) in the world from the high-level planner and use the cascade PID-based control as follows:
\begin{equation}
\begin{split}
\bm{v}_{err} &= K_{pp}(\bm{p}_d-\bm{p}) \\
\bm{v}_d &= \bm{v}_d +\bm{v}_{err}\\
\bm{a}_{err} &= K_{vp}(\bm{v}_d-\bm{v}) +K_{vi}\int{(\bm{v}_d-\bm{v}) dt}+K_{vd}(\dot{\bm{v}}_d-\bm{v}) 
\label{con:equ16}
\end{split}
\end{equation}
where $\bm{a}_{err}$ represents the corresponding outputs for PID-based output, with a control frequency of 100 Hz. $K_{pp},K_{vp},K_{vi},K_{vd}$ are the parameters of the controller.
With (~\ref{con:equ22}) and (~\ref{con:equ23}), the actual acceleration of the quadrotor is: 
\begin{equation}
\begin{split}
\bm{a}_c = \ddot{\bm{p}}_{d} + (\bm{f}^{high}+\bm{f}^{low})/m_B,
\label{con:equ17}
\end{split}
\end{equation}
where $\ddot{\bm{p}}_{d}$ represents the desired acceleration of the quadrotor obtained from the trajectory.

\subsubsection{Determination of Thrust and Angular Velocity}
With the actual desired acceleration, yaw($\phi$), and yaw angular rate, we determine the thrust and attitude using differential flatness, incorporating rotor drag as described in \cite{faessler2017differential}:
\vspace{-10pt}
\begin{equation}
\begin{split}
\bm{x}_C&=\begin{bmatrix}\cos(\psi)&\sin(\psi)&0\end{bmatrix}^T\\\bm{y}_C&=\begin{bmatrix}-\sin(\psi)&\cos(\psi)&0\end{bmatrix}^T\\\bm{z}_{B,des}&=\frac{\bm{a}_{des}}{\|\bm{a}_{des}\|}\\
\bm{x}_{B,des}&=\frac{\bm{y}_{C}\times\bm{z}_{B,des}}{\|\bm{y}_{C}\times\bm{z}_{B,des}\|}\\
\bm{y}_{B,des}&=\bm{z}_{B,des}\times\bm{x}_{B,des}.
\label{con:equ18}
\end{split}
\end{equation}
 
To achieve more robust and precise control, we convert the desired attitude into the corresponding thrust and angular velocity, which are then sent to the underlying PX4\footnote{\url{https://github.com/PX4/PX4-Autopilot}} controller:
\begin{equation}
\begin{split}
T_d=\mathbf{z}_\mathrm{B}^T\left(\mathbf{a}+g\mathbf{z}_\mathrm{W}+d_\mathrm{z}\mathbf{v}\right).
\label{con:equ19}
\end{split}
\end{equation}

 Due to space constraints, we don't give the angular velocity formulation. If readers interested in further details are strongly encouraged to refer to \cite{faessler2017differential}.

\subsection{Delta-Arm Control}
\label{manipulator control}

The delta-arm actuators are the Dynamixel servos, which are capable of providing position and velocity information, thereby enabling velocity control. As depicted in Fig~\ref{fig:3}, the higher-level planner publishes the desired position ($\bm{p}_d^e$) and the desired velocity ($\bm{v}_d^e$). To achieve smooth tracking of the desired trajectory for the end-effector, we utilize PD-based control:
\begin{equation}
\begin{split}
\bm{v}_{err} &= K_{p}(\bm{p}_d^e-\bm{p}^e) +K_{d}(\dot{\bm{p}}_d^e-\bm{p}_d^e) 
\\
\bm{v}_c^e &= \bm{v}_d^e+\bm{v}_{err}.
\label{con:equ24}
\end{split}
\end{equation}
where $\bm{v}_c^e$ represents the final desired velocity of the end-effector. We use the velocity inverse kinematics equation \eqref{con:velocity con11} to convert the $\bm{v}_c^e$ into the desired angular velocity $\bm{\omega}_d^e$. This desired angular velocity is then sent to the servo actuator, which actively tracks the commanded velocity and provides feedback in the form of $\bm{p}^e$ and $\bm{v}^e$ information at a rate of 100 Hz. This feedback loop enhances the position control of the delta-arm end-effector.

\section{EXPERIMENTS}
\label{SECTION V}
We evaluate the performance of the proposed control framework through adequate real-world experiments. In this section, we aim to evaluate our control framework through three carefully selected experiments. \expandafter{\romannumeral1}) 
Conduct an experiment where the end-effector moves rapidly while carrying a payload, and the quadrotor remains stabilized at the origin~(\ref{Disturbance Rejection}).
\expandafter{\romannumeral2}) 
Perform an experiment where the aerial manipulator tracks a trajectory under wind conditions, and the end-effector compensates for the tracking errors~(\ref{Trajectory compensation}).
\expandafter{\romannumeral3})
Execute an experiment where the quadrotor and the manipulator track different trajectories while keeping the end-effector stationary at the origin~(\ref{end-effector Stablization }).

\subsection{Disturbance Rejection}
\label{Disturbance Rejection}

\begin{figure}
    \centering
    \includegraphics[width=1\linewidth]{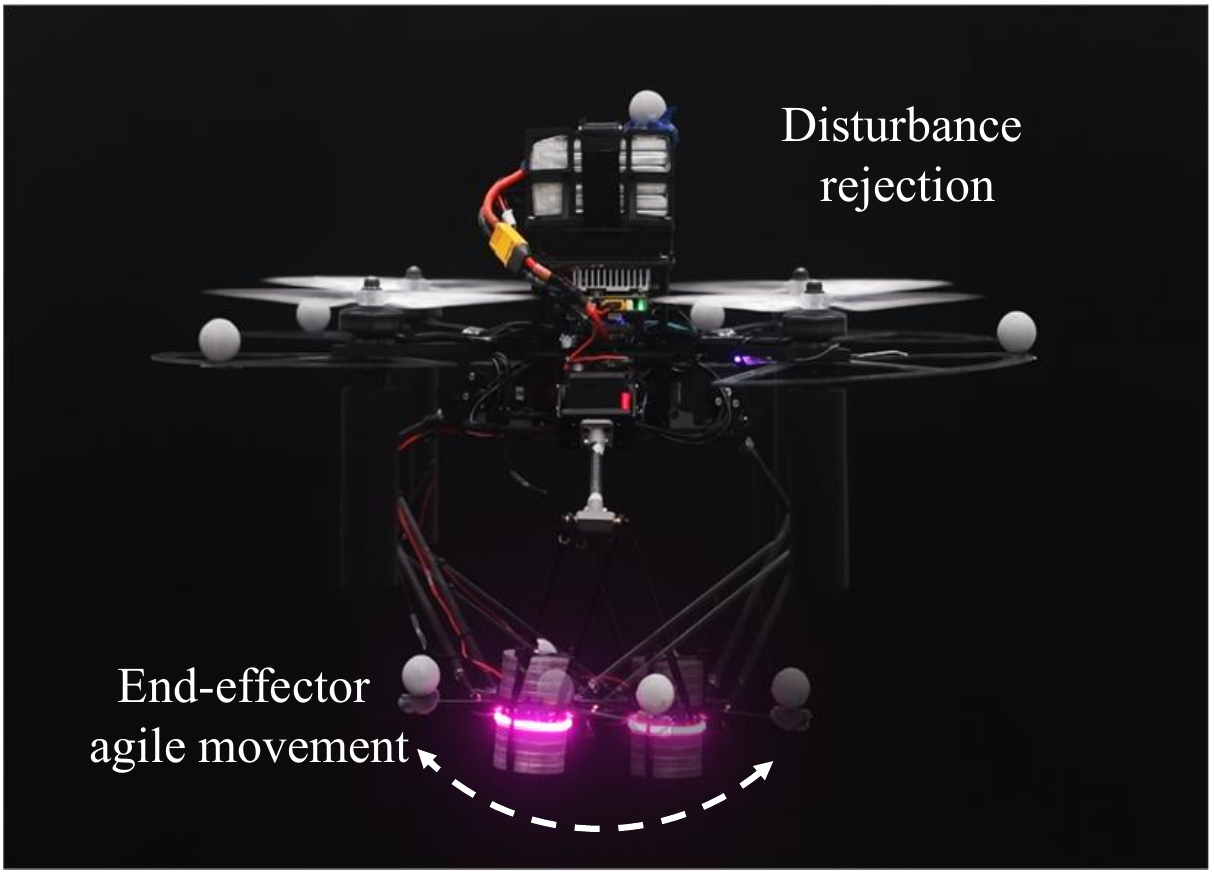}
    \caption{Snapshot of disturbance rejection. The end-effector’s trajectory is depicted as a dotted line. In this test, the end-effector performed agile movements, carrying a 400~g payload, while the quadrotor stablize to the origin to prove its robustness. With our control framework, it will have 80\% performance improvement compared to the PX4.}
    \label{disturbance rejection picture}
    \vspace{-20pt}
\end{figure}

In this experiment, we conducted three ablation studies, each repeated three times. The mean of the results was taken as the final outcome. The first group uses the PX4 algorithm, which is widely employed in aerial manipulators \cite{soria20193d, sun2021switchable}, without any improvements, like NDOB, or high-frequency compensation. The second group incorporates only the NDOB without high-frequency compensation. The third group utilizes the method we proposed. Fig.~\ref{disturbance rejection picture} presents the snapshot obtained from the experiment. As shown in Table~\ref{disturbance rejection table} (all units in meters), for each condition, the Root Mean Square Error (RMSE) during UAV stabilization and the maximum deviation from the original position (Max Error) were measured. 
Compared to the PX4 method, the NDOB method shows significant improvements under all conditions. For the condition of 10 cm/s velocity and 400 g mass, the RMSE and Max Error are reduced by 62.14\% and 52.17\%, respectively. Furthermore, under high-speed conditions (10 cm/s velocity and 400 g mass), our method demonstrates further improvements compared to the NDOB method. The results show our proposed method significantly enhances the agility and robustness of aerial manipulator.

\subsection{Trajectory Compensation}
\label{Trajectory compensation}

\begin{figure}
    \centering
    \includegraphics[width=1\linewidth]{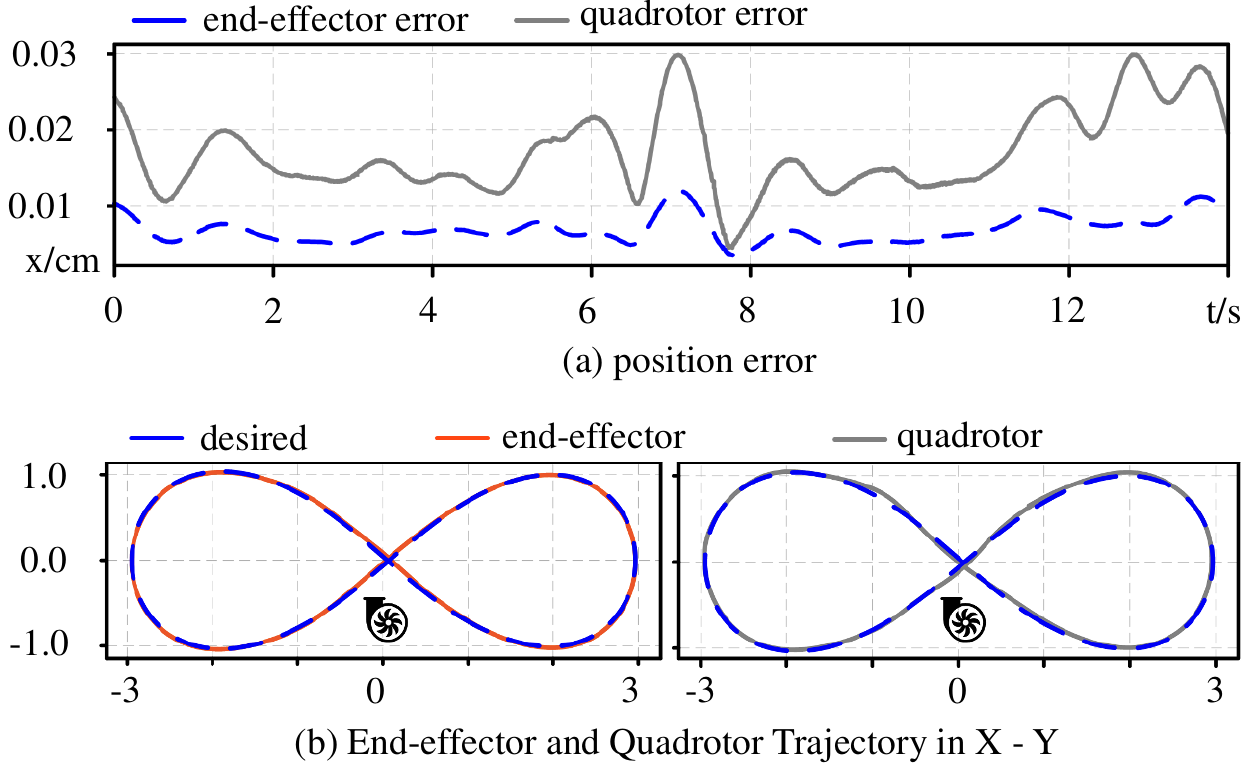}
    \caption{ (a) shows the end-effector and quadrotor position error. (b) shows the end-effector and quadrotor tracking trajectory in the X-Y plane.}
    \label{compensate trajectory picture}
    \vspace{-50pt}
\end{figure}
In the second experiment, the control objective is for the end-effector to accurately track a complex desired trajectory, specifically the figure-eight shape generated by \cite{wang2022geometrically}. The end-effector will compensate for the tracking error caused by the quadrotor with a 200 g payload. As the end-effector operates in three-dimensional space with position, we deliberately limit the quadrotor's velocity (maximum 1.5 m/s) to ensure stable attitude control, making it easier for the end-effector to compensate for errors.

\begin{figure}
    \centering
    \includegraphics[width=1\linewidth]
    {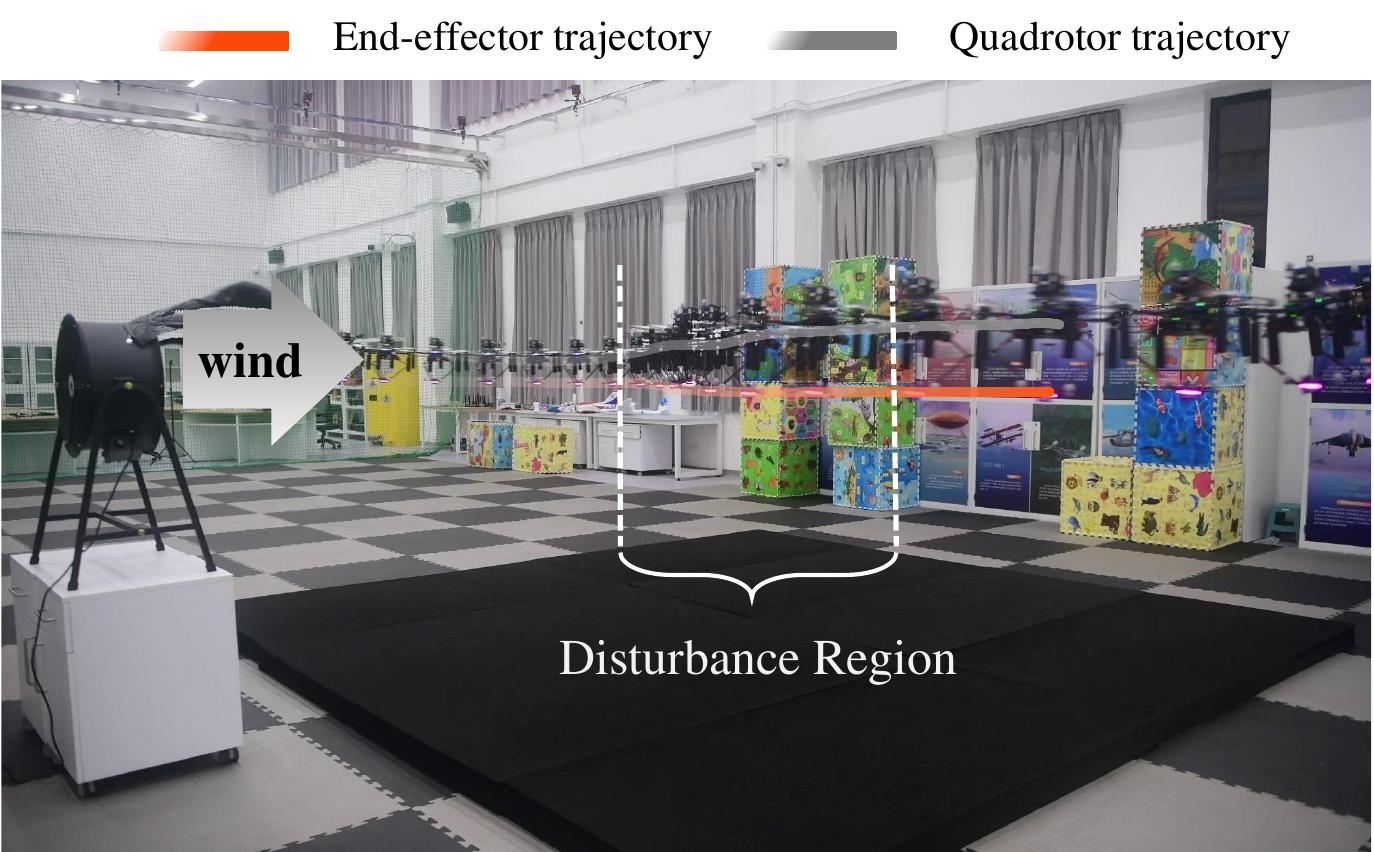}
    \caption{  Real-world trajectory tracking experiments. Snapshots of trajectory tracking tests under wind environment.}
    \label{compensate trajectory real picture}
    \vspace{-20pt}
\end{figure}

The objective of this experiment is to validate the AM's capability to track trajectories, even in the presence of external wind disturbances. 
The result of experiment is shown in Fig.~\ref{compensate trajectory picture} (a), illustrating the position tracking comparison with a 200 g payload. Fig.~\ref{compensate trajectory picture} (b) displays the AM tracking trajectory in the X-Y plane. The left side shows the end-effector tracking trajectory in the X-Y plane, while the right side shows the quadrotor tracking trajectory in X-Y plane. It is evident that the end-effector exhibits superior tracking performance compared to the quadrotor. As observed, the RMSE for the quadrotor is 17 mm, while the end-effector achieves an RMSE of 6 mm, marking a notable 64\% enhancement in performance. The experiment picture is shown in Fig.~\ref{compensate trajectory real picture}.
\vspace{-10pt}
\begin{table}[h]
\centering
\caption{Control Performance Comparison}\label{disturbance rejection table}
\resizebox{\columnwidth}{!}{
\begin{tabular}{lcccccc}
\toprule & \multicolumn{2}{c}{5cm/s 200g} & \multicolumn{2}{c}{5cm/s 400g} & \multicolumn{2}{c}{10cm/s 400g}  \\
\cmidrule(r){2-3} \cmidrule(r){4-5} \cmidrule(r){6-7}  
& RMSE & Max Err. & RMSE & Max Err. & RMSE & Max Err. \\
\midrule
PX4  & 0.09 & 0.12 & 0.12 & 0.17 & 0.14 & 0.23 \\
NDOB & 0.03 & 0.028 & 0.042 & 0.067 & 0.053 & 0.11 \\
Ours & \textbf{0.013} & 0.029 & \textbf{0.021} & 0.044 & \textbf{0.029} & 0.057\\
\bottomrule
\end{tabular}
}
\vspace{-10pt}
\end{table}

\subsection{End-effector Stablization }
\label{end-effector Stablization }
The third experiment is to stabilize the end-effector at a fixed position while the quadrotor tracks along a circle trajectory with a diameter of 0.12 m (see Fig. \ref{fig:1}). This scenario is crucial because practical tasks often involve assigning different trajectories for the quadcopter base and the end-effector.
For example, the end-effector may need to track a specified trajectory to complete an interactive task, while the quadcopter base may need to navigate to avoid obstacles or observe the manipulated target from various angles using onboard cameras. 
To achieve this, the aim is to maintain $\bm{p}^e$ constant and minimize $\dot{\bm{v}}^e$ to zero.
The result is shown in  
table~\ref{End-effector Stablization table} (all units in meters).  For each condition, the RMSE of the end-effector's position relative to its initial position during stabilization, and the standard deviation (std) from the original position was measured.
Compared to the state-of-the-art ESO method \cite{cao2023eso}, our approach achieves velocities for both the quadrotor base and the end-effector that are five times faster. Furthermore, our RMSE is 50\% lower, measuring only 6mm. 
\begin{table}[h]
\centering
\caption{End-effector Stablization Performance Comparison}\label{End-effector Stablization table}
\resizebox{\columnwidth}{!}{
\begin{tabular}{lccccccc}
\toprule & \multicolumn{2}{c}{1cm/s 0.08m} & \multicolumn{2}{c}{1cm/s 0.12m} & \multicolumn{2}{c}{5cm/s 0.12m}  \\
\cmidrule(r){2-3} \cmidrule(r){4-5} \cmidrule(r){6-7}  
& RMSE & STD. & RMSE & STD. & RMSE & STD. \\
\midrule
PX4 & 0.019 & 0.009 & 0.030 & 0.009 & 0.048 & 0.012 \\
ESO\cite{cao2023eso} & - & - & 0.011 & 0.004 & - & - \\
Ours & \textbf{0.005} & 0.004 & \textbf{0.005} & 0.004 & \textbf{0.006} & 0.005\\
\bottomrule
\end{tabular}
\vspace{-5pt}
}
\end{table}

\section{CONCLUSION AND FUTURE WORK}
\label{SECTION VI}

This paper presents a robust control framework for achieving agile end-effector movements in an aerial manipulator, employing an NDOB with high-frequency compensation for end-effector dynamics. Our approach enhances the agility and smoothness of the aerial manipulator's movements. Experimental results demonstrate the efficacy of our method in various challenging scenarios, highlighting significant improvements in tracking performance and disturbance rejection. With a 400 g payload, the quadrotor maintains stable control during agile movements of the end-effector, achieving an RMSE of 0.029 m. When the quadrotor and manipulator track separate trajectories, the end-effector achieves a control error of just 6 mm, operating at a speed five times faster than the current state-of-the-art (SOTA). Future research will focus on refining the applications of the aerial manipulator, particularly when using a delta-arm end-effector that operates in parallel with the quadrotor.
\bibliography{IROS_2024_cxd} 

\end{document}